\documentclass[lettersize,journal]{IEEEtran}
\usepackage{hyperref}
\pdfoutput=1
\DeclareUnicodeCharacter{2217}{\'{*}}

\usepackage{soul}
\soulregister\cite7 
\soulregister\citep7 
\soulregister\citet7 
\soulregister\ref7 
\soulregister\pageref7 
\usepackage{mathrsfs}
\usepackage[noend]{algpseudocode}
\usepackage{multirow}

\usepackage{algorithmicx,algorithm}
\usepackage{bbm}
\usepackage{graphicx}
\usepackage{amsmath}
\usepackage{amssymb}
\usepackage{multirow}
\usepackage{helvet}
\usepackage{courier}
\usepackage{float}
\usepackage{color,soul,xcolor}

\usepackage{amsfonts}
\usepackage{textcomp}
\usepackage{hyperref}

\usepackage{subfigure}
\usepackage{hyperref}
\usepackage{times}
\usepackage{epsfig}

\usepackage{array}
\usepackage{textcomp}
\usepackage{stfloats}
\usepackage{url}
\usepackage{verbatim}
\usepackage{arydshln}
\usepackage{bbding}
\usepackage{wrapfig}
\usepackage{caption}
\usepackage{booktabs}
\usepackage{colortbl}
\usepackage{newfloat}
\usepackage{listings}
\usepackage{bm}

\usepackage{makecell}
\usepackage{pifont}
\usepackage{tabularx}

\newcommand{\redxmark}{\textcolor{red}{\ding{55}}}
\newcommand{\bluecheckmark}{\textcolor{blue}{\ding{51}}}
\usepackage{tikz}
\usepackage{tabularray}
\definecolor{MineShaft}{rgb}{0.2,0.2,0.2}

\begin{document}

\title{Advances in Embodied Navigation \\Using Large Language Models: A Survey}

\author{Jinzhou Lin,\thanks{Jinzhou Lin and Han Gao contribute equally.} 
Han Gao, 
Xuxiang Feng,
Rongtao Xu$^{\dag}$,
Changwei Wang,
Dong An,
Jie Zhou,
Man Zhang,
Li Guo,

Xiaoqiang Teng,
Shibiao Xu
$^{\dag}$,~\IEEEmembership{Member,~IEEE,} \thanks{Rongtao Xu and Shibiao Xu are the corresponding authors (xurongtao2019@ia.ac.cn; shibiaoxu@bupt.edu.cn).}\thanks{Jinzhou Lin, Han Gao,Jie Zhou, Man zhang, Li Guo and Shibiao Xu are with School of Artificial Intelligence, Beijing University of Posts and Telecommunications, China.}
\thanks{Xuxiang Feng is with the Aerospace Information Research Institute, Chinese Academy of Science.}\thanks{Rongtao Xu and Dong An are with the State Key Laboratory of Multimodal Artificial Intelligence Systems, Institute of Automation, Chinese Academy of Sciences, China.}\thanks{Changwei Wang is with Key Laboratory of Computing Power Network and Information Security, Ministry of Education, Shandong Computer Science Center (National Supercomputer Center in Jinan), Qilu University of Technology (Shandong Academy of Sciences), also with Shandong Provincial Key Laboratory of Computing Power Internet and Service Computing, Shandong Fundamental Research Center for Computer Science, Jinan, 250014, China}\thanks{Xiaoqiang Teng is with the School of Computer and Artificial Intelligence, Beijing Technology and Business Unicersitu, China.}\thanks{This work is supported by the National Natural Science Foundation of China, No.62302052}
}

\markboth{ IEEE TRANSACTIONS ON SYSTEMS, MAN, AND CYBERNETICS: SYSTEMS}%
{Jinzhou Lin and Han Gao \MakeLowercase{\textit{et al.}}: Advances in Embodied Navigation Using Large Language Models: A Survey}
\maketitle

\begin{abstract}
Large Language Models like GPT have gained significant attention for their potential in various practical applications. One notable area of focus is their integration with Embodied Intelligence, particularly in navigation tasks, which require deep environmental understanding and rapid, accurate decision-making. LLMs enhance embodied intelligence systems by providing advanced environmental perception and decision-making support, leveraging their language and image-processing capabilities. This article reviews the symbiosis between LLMs and embodied intelligence, focusing on navigation, and evaluates current models, research methodologies, and datasets. Additionally, it discusses the role of LLMs in the field and forecasts future research directions. A comprehensive list of studies is available at \href{https://github.com/Rongtao-Xu/Awesome-LLM-EN}{https://github.com/Rongtao-Xu/Awesome-LLM-EN}.

\end{abstract}

\begin{IEEEkeywords}
Large Language Models, Embodied Intelligence, Navigation.
\end{IEEEkeywords}


\section{Introduction}

\IEEEPARstart{T}he development of Large Language Models (LLMs) in the field of navigation is rapidly evolving and holds significant potential for advancing natural language processing and machine learning. Notably, LLMs have already achieved remarkable success in few-shot planning, enabling effective planning and decision-making for new tasks with minimal or no sample data. However, alongside these achievements, numerous technical and theoretical challenges persist. These challenges include the integration of text, images, and other sensor data simultaneously, the reduction of latency for real-time applications, and the enhancement of training efficiency without sacrificing performance.

To address these challenges, researchers have employed a diverse array of methods such as machine learning~\cite{ren2020advising}, deep learning~\cite{xu2024local}\cite{wang2024efrnet} and evolutionary algorithms~\cite{20230721VoxPoser}~\cite{stolfi2024evolutionary}, as well as the addition of other vision-language models to enhance the multimodal data fusion capabilities of LLMs\cite{gonzalez2022detection}. These methodologies aim to develop navigation agents capable of learning from experience and continually improving their performance\cite{zhao2023autonomous}. In this paper, we review existing models that have employed LLM-based agents for navigation tasks. These studies have demonstrated that LLM-based agents perform excellently in path planning, environmental understanding, and dynamic adjustment tasks. The success of these studies underscores the utility of LLM-based agents as valuable tools for navigation.

This paper primarily explores the usage of LLMs in navigation models. By delving into these technologies, we provide comprehensive knowledge and resources for researchers and practitioners. Compared to other related reviews, this paper focuses specifically on the application of large language models in the field of navigation, analyzing the strengths and weaknesses of existing LLM-based navigation models, and comparing them with non-LLM-based Vision-and-Language Navigation (VLN) models~\cite{cui2024survey}~\cite{zhang2022survey}~\cite{evaluation}. Additionally, we review common datasets in the navigation field, compare these datasets, and highlight the current issues with them. Our contributions are as follows:
\begin{enumerate}
  \item \textbf{Comprehensive Review and Comparative Analysis}: The article provides a comprehensive review of the application of LLMs  in the navigation field, covering multiple LLM-based navigation models and comparing them with non-LLM-based models. This integrated analysis helps readers fully understand current research developments and highlights the advantages and disadvantages of LLMs in navigation.
  \item \textbf{Dataset Analysis}: The article provides a thorough analysis of commonly used datasets in the navigation field, discussing their applicability, issues, and limitations. This detailed analysis helps researchers select appropriate datasets and offers practical advice on their application, thereby enhancing the practical relevance of the article.
  \item \textbf{Challenges and Future Directions}: The article not only showcases current research advancements but also delves into the challenges and issues faced by existing methods in practical applications. By identifying these challenges, the article points out directions for future research and highlights innovative application scenarios and technological trends of LLMs in the navigation field.
\end{enumerate}

\section{Background}
\subsection{Large Language Models}
\sloppy
    \begin{figure*}[t]
      \centering
      \includegraphics[width=\linewidth]{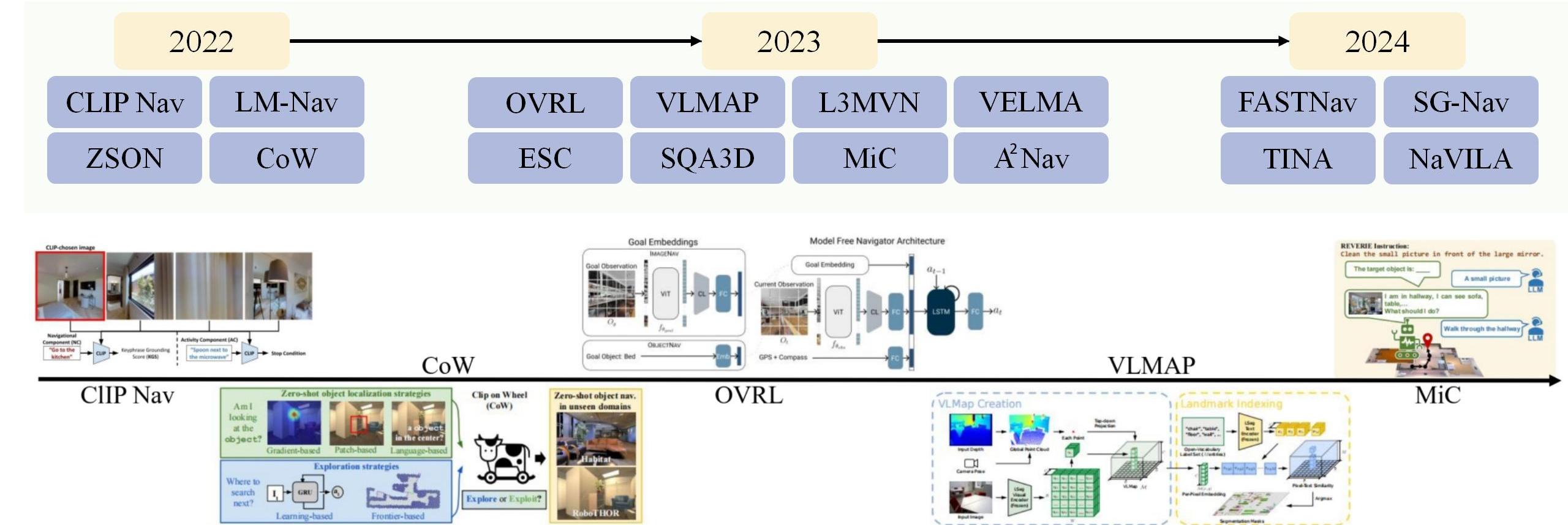}
      \caption{This presentation exhibit a temporal map depicting the works of embodied navigation from 2022 to 2024, and we selected 5 typical works to showcase their corresponding framework diagrams. The map illustrates the evolution of major works, offering valuable insights into the advancement of Embodied Agents.
}
          \label{fig:BenchmarksTimeMap}
    \end{figure*}
The advent of LLMs marks a pivotal achievement in the domains of Natural Language Processing (NLP) and Machine Learning. LLMs are sophisticated machine learning models specifically engineered to process and generate natural language, typically trained on vast text corpora to capture syntactic structures, inter-word relationships, and contextual subtleties~\cite{naveed2023comprehensive}\cite{li2021creating}. These models predominantly utilize the Transformer architecture, which incorporates a self-attention mechanism to adeptly manage long-range dependencies, making them suitable for a wide array of tasks, such as natural language generation, translation, summarization, and question answering.

The Transformer architecture, introduced by Google in 2017~\cite{vaswani2017attention}, eschews recurrent and convolutional layers in favor of a self-attention mechanism to capture sequence dependencies, thereby dramatically enhancing processing speed over traditional RNNs and LSTMs. Following this, models such as ViT~\cite{ViT} and CLIP~\cite{radford2021learning} expanded the application of Transformer-based models from NLP to visual processing tasks. BERT~\cite{devlin2018bert}, introduced by Google in 2018, leverages bidirectional encoders to capture both antecedent and subsequent contextual information, while the GPT series, known for its focus on text generation, demonstrates "zero-shot learning" capabilities, enabling models to execute tasks without task-specific training~\cite{gpt1}\cite{gpt2}\cite{gpt3}.

LLMs have found widespread application across various fields, including NLP, autonomous driving, and robotics task planning. In particular, within autonomous driving and robotics task planning, LLMs facilitate the understanding of natural language commands and the generation of precise action plans, thereby enhancing both user experience and operational efficiency~\cite{cui2024survey}\cite{zeng2023large}\cite{wang2024large}. However, despite their exceptional adaptability and capacity for real-time data processing, LLMs demand substantial computational resources and rely heavily on high-quality, extensive datasets~\cite{20230712VELMA}.

In conclusion, LLMs exhibit immense potential across a broad range of domains, yet they are confronted with challenges related to data diversity, granular navigation, spatial reasoning, and interaction adaptability. Future advancements are poised to focus on multimodal integration and optimization strategies~\cite{yu2023merlin}\cite{wang2022towards}.

\subsection{Embodied Intelligence}
  Embodied Intelligence is an emerging field that explores intelligent agents interacting with their environment, with the premise that intelligence arises from this interaction rather than abstract computation. This field integrates neuroscience, psychology, robotics, and AI to develop models and algorithms for simulating intelligent behavior~\cite{chen2024semnav}.

Embodied navigation refers to autonomous movement through perception, decision-making, and action within a physical environment\cite{ren2024infiniteworld}. It requires intelligent agents to understand their surroundings through sensory systems and adapt navigation behaviors based on physical attributes such as size, weight, and mobility~\cite{zhao2024information}. Challenges include dynamic obstacles, terrain variations, limited sensory data, and the need for real-time decision-making~\cite{zhong2024casit}.

Recent advancements include integrating LLMs with intelligent agents. These models leverage the NLP capabilities of LLMs to translate human instructions into formats interpretable by agents. Researchers are moving beyond text inputs to incorporate images, point clouds~\cite{10375828}\cite{10274671}, and voice prompts, aligning with the concept of multimodal algorithms\cite{li2024multi}. Multimodal integration allows agents to gather comprehensive environmental data through visual, auditory, and tactile cues, improving navigation in real-world scenarios~\cite{xu2024deffusion}~\cite{shi2023intelligent}.

Models like DALL-E and Vision Transformers~\cite{ViT} enhance multimodal interaction by processing images and combining them with text for better user interaction. These systems provide decision support, enabling robots to understand user needs and enhance the navigation experience.

In embodied navigation, agents' control is divided into high-level and low-level aspects. High-level controls focus on task scheduling and strategy development through techniques like reinforcement learning~\cite{guo2023optimal}, while low-level controls manage direct operations such as speed and position~\cite{Beranek}. These methodologies enable applications in terrain recognition~\cite{9183954}, machinery lifespan prediction~\cite{10053378}, and gaze mechanisms~\cite{9916584}, with significant advancements in agent robustness and generalization.

In conclusion, the field of embodied navigation is evolving from reinforcement learning to LLM integration and multimodal approaches~\cite{xu2024mrftrans}. The integration of LLMs and other algorithms continues to enhance intelligent agent capabilities, with a focus on improving adaptability and performance in real-world applications.

    \begin{figure*}[ht]
      \centering
      \includegraphics[width=\linewidth]{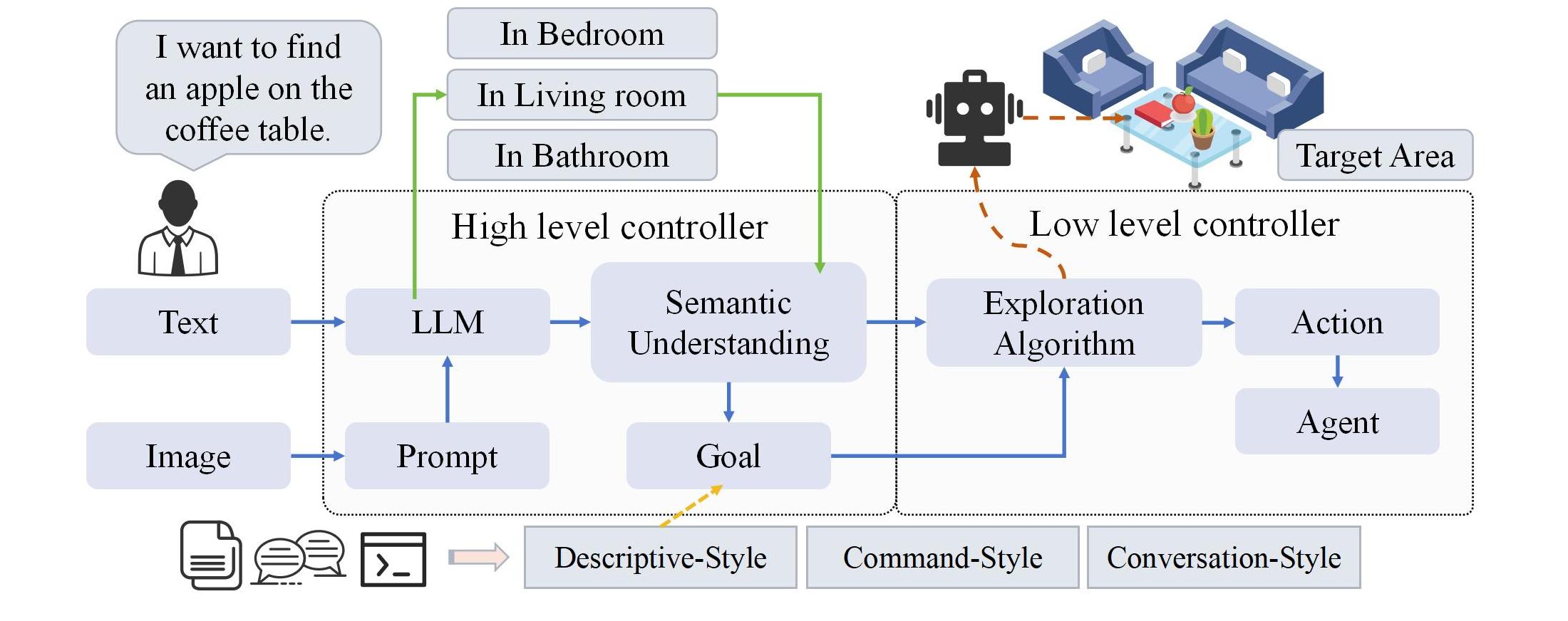}
      \caption{The first type utilizes LLMs to analyze incoming visual or textual data to extract goal-relevant information, upon which exploration policies subsequently generate appropriate actions to guide agent movement. LLMs, by acquiring information from text and other visual models processed through images, perform semantic understanding rather than planning. They extract key information sequences such as targets and locations, and hand them over to exploration algorithms. The exploration algorithms generate actions to guide agents in navigation to complete tasks.
}
      \label{fig:SemanticPic}
    \end{figure*}
\section{LLMs in Embodied Navigation}
Large language models often play two roles in navigation: either as planning models or as information-providing models, as described in this section under "LLMs for Grounded Language Understanding" and "LLMs for Few-Shot Planning."
\subsection{LLMs for Grounded Language Understanding}
  Grounded Language Understanding aims to reconcile the abstract symbols processed by language models with concrete entities, actions, or states in the physical or simulated world. This is pivotal for applications requiring real-world interaction, such as robotic control, natural language interfaces, or advanced research in Embodied Intelligence~\cite{xu2025a0}.

  Within this context, LLMs like GPT and BERT can integrate with sensors, databases, or simulated environments to generate and interpret language applicable to real-world scenarios. In a robotic setting, for instance, an LLM could interpret sensor data, process natural language directives, and issue control signals to the robot, thereby bridging high-level language and low-level actions.

  LLMs' application in grounded language understanding is an evolving research area, intersecting with disciplines such as robotics, computer vision, and human-computer interaction. Techniques like Few-Shot Learning and Transfer Learning are commonly used to adapt these pre-trained models to specialized grounding tasks with minimal additional training.

  Although LLMs have advanced significantly in NLP, they primarily excel in text generation or classification and seldom engage with tangible entities or real-world situations. Their limitations include a lack of foundational knowledge, impairing their ability to process interrelated concepts and function in interactive settings (Mahowald et al.~\cite{mahowald2023dissociating}\cite{cebollada2021state}). To mitigate these issues, researchers have employed various strategies, such as 1) fine-tuning pre-trained models, 2) implementing reinforcement learning algorithms for decision-making in complex environments, and 3) devising specialized architectures for multimodal learning.

  Despite their immense potential and practical applications, LLMs face numerous technical and theoretical challenges in grounded language understanding. These include the integration of text, images, and sensor data, latency reduction for real-time applications, and maintaining training efficiency without sacrificing performance. With ongoing research and emerging technologies, significant advancements in this domain are anticipated.
    \begin{figure*}[ht]
      \centering
      \includegraphics[width=\linewidth]{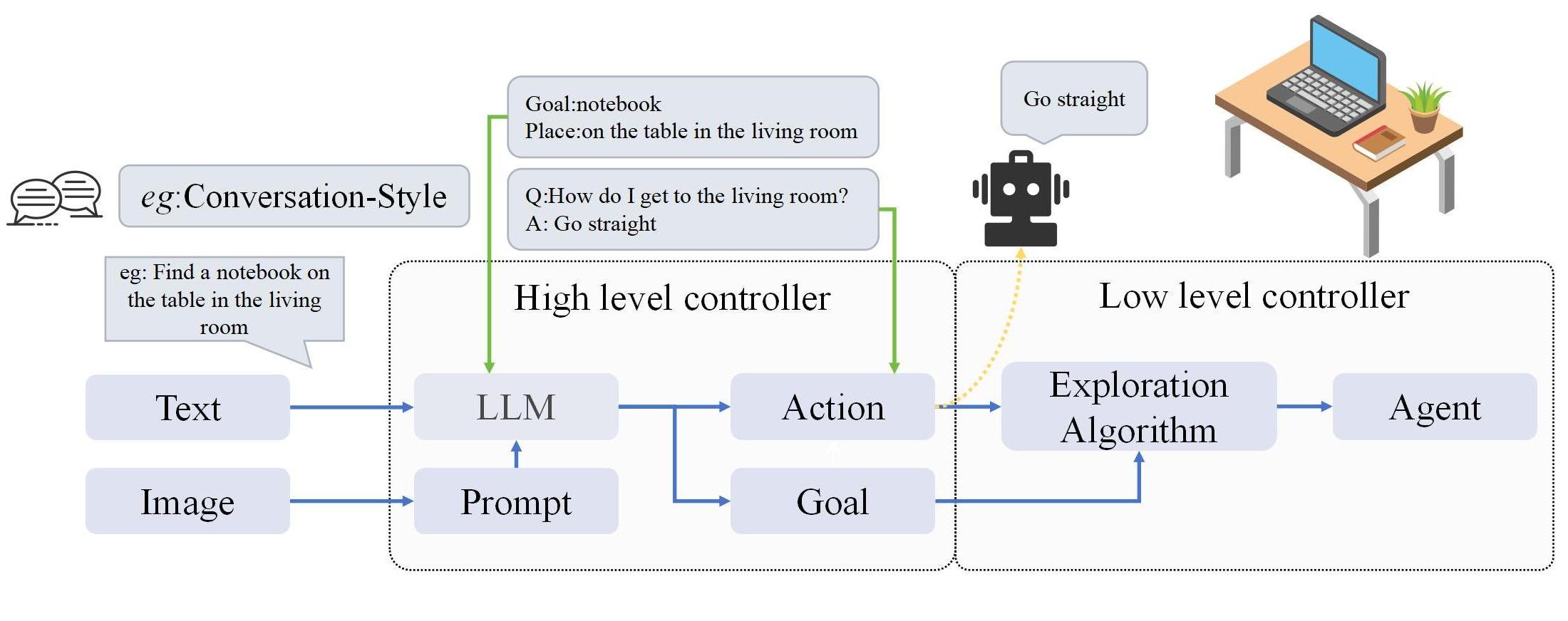}
      \caption{The second type employs LLMs as planners that directly generate actions, thereby leveraging exploration policies to control agents. LLMs, by acquiring information from text and other visual models processed through images, perform planning (using a dialogue format as an example here) and hand the actions over to exploration algorithms and agents for navigation to complete tasks.
}
      \label{fig:PlanningPic}
    \end{figure*}
\subsection{LLMs for Few-Shot Planning}
 LLMs excel in zero-shot and few-shot learning, allowing for efficient generalization across tasks without the need for task-specific training or limited examples. Employing LLMs for few-shot planning often involves processing natural language queries, enabling the models to generate actionable plans or steps, thus proving invaluable in zero-shot navigation tasks~\cite{liang2025structured}.

  As discussed in research by Zhou et al. in ESC~\cite{20230527ESC} and Song et al. in LLM-Planner~\cite{llmplanner}\cite{ntakolia2023autonomous}, the effectiveness of navigation tasks is closely related to semantic scene understanding, which relies heavily on Few-Shot Planning. Planners incorporating LLMs are increasingly vital for enabling embodied agents to execute complex tasks in visually rich environments based on natural language directives. The performance of such planners has a direct bearing on the task's overall success. The role of LLMs as planners in embodied tasks is expanding, thanks in part to the emergence of methodologies like LLM-Planner~\cite{llmplanner}, CoT (Chain-of-Thought)\cite{CoT}, and LLM-DP\cite{llmdp}. These methods complement the already respectable performance of traditional symbolic planners like the Fast-Forward planner~\cite{ff} and BFS(f) planner~\cite{BFS(f)} in current research.

\subsection{Embodied Navigation}
   In the field of embodied navigation, state-of-the-art models enhance their performance and practicality through various methods. Firstly, these models achieve multimodal integration~\cite{xu2024mrftrans}\cite{yin2024autonomous}, combining visual, linguistic, and auditory data, using LLMs to process complex language instructions in conjunction with visual systems to accurately interpret and respond to the environment~\cite{20230911SayNav}\cite{lee2024lsmcl}. Secondly, they enhance understanding of environmental context and decision-making through deep learning algorithms~\cite{xu2024local}\cite{samarakoon2022metaheuristic}, leveraging the advanced language comprehension capabilities of LLMs to better grasp task requirements and navigate environmental obstacles~\cite{20230529Navgpt}\cite{naghizadeh2020gnm}. Additionally, models with long-term memory mechanisms can remember past navigation experiences and environmental features, aiding effective path planning in complex or changing environments~\cite{20230529Navgpt}~\cite{qiao2023march}\cite{meysami2024efficient}. These models also possess the ability to interact effectively with users, processing natural language inputs to communicate in a more natural and human-like manner\cite{zhao2023autonomous}. Lastly, the adoption of reinforcement learning methods allows models to self-optimize and learn through continuous interaction with the environment, continually adjusting behaviors to adapt to new tasks or environmental changes~\cite{wang2023seamless}. By integrating these technologies, state-of-the-art embodied navigation models not only improve navigation efficiency but also enhance their ability to cope with complex environments, achieving highly intelligent and adaptable navigation capabilities~\cite{zhang2022survey}\cite{aboyeji2024covariance}.

   Embodied navigation is crucial for enabling robots to interact effectively with their environments, involving complex tasks like sensory perception, decision-making, and motion planning. Recent research advances have focused on improving path planning algorithms for dynamic and complex environments~\cite{song2020neural}. \cite{li2024dynamic} introduces adaptive dynamic programming for real-time path adjustments in mobile robots, while \cite{parhi2023humanoid} applies a memory-based gravity search algorithm for humanoid robot navigation. \cite{madridano2021trajectory} reviews trajectory planning methods for multi-robot systems, and \cite{jeyalakshmi2024agile} develops hybrid optimization techniques for autonomous car navigation. 
 \subsubsection{LLM-based Model}

 Generally, LLM-based navigation models employ approaches that fall into two categories, as illustrated in Fig.\ref{fig:PlanningPic} and Fig.\ref{fig:SemanticPic}. The first category uses LLMs as planners, directly generating actions and leveraging exploration policies to guide agents. The second category employs LLMs to scrutinize incoming visual or textual data to isolate goal-relevant information, based on which exploration policies subsequently produce suitable actions for agent navigation. An architectural diagram delineating these approaches is presented below. This section endeavors to examine the nuanced features and distinctions among these works, furnishing a comprehensive temporal map of the works under discussion in this chapter, as depicted in Fig.\ref{fig:BenchmarksTimeMap}.

\begin{table*}[ht]
\center

\caption{Classification of the works described in this section.}
\begin{tabular}{ccccccc}
\multicolumn{1}{c|}{Categories}         & \multicolumn{6}{c}{Works}                                                                                                                              \\ \hline
\multirow{2}{*}{Planning}               & \multirow{2}{*}{CLIP-Nav} & \multirow{2}{*}{NavGPT} & \multirow{2}{*}{VELMA} & \multirow{2}{*}{MiC}    &\multirow{2}{*}{SayNav}  &  \\
                                        &                           &                         &                        &                        &                      &                         \\ \hline
\multirow{2}{*}{Semantic Understanding} & \multirow{2}{*}{LM-Nav}   & \multirow{2}{*}{BEVBert}  & \multirow{2}{*}{ESC}   & \multirow{2}{*}{SQA3D} & \multirow{2}{*}{}    & \multirow{2}{*}{}       \\
                                        &                           &                         &                        &                        &                      &                         \\ \hline
\multicolumn{1}{l}{}                    & \multicolumn{1}{l}{}      & \multicolumn{1}{l}{}    & \multicolumn{1}{l}{}   & \multicolumn{1}{l}{}   & \multicolumn{1}{l}{} & \multicolumn{1}{l}{}   
\end{tabular}
\end{table*}
  
    a) \textbf{LM-Nav}~\cite{20220726LMNAV} is a navigation architecture designed for robots that leverages pre-existing models specialized in language, vision, and action to enable sophisticated interactions with robots via natural language commands. Remarkably, the system eliminates the need for costly supervision and fine-tuning, relying solely on pre-trained models for navigation, image-langu-age correlation, and language modeling. The architecture of LM-Nav consists of three integrated, pre-trained models to ensure precise instruction execution in complex, real-world scenarios. Specifically, the system employs GPT-3 as the LLM, tasked with decoding verbal instructions into a series of textual landmarks. Concurrently, CLIP serves as the VLM, anchoring these textual landmarks to a topological map. Lastly, the Vision-Action Model (VAM) is a self-supervised robotic control model, responsible for utilizing visual data and executing physical actions based on plans synthesized by the LLM and VLM.

    Implemented on a real-world mobile robot, LM-Nav has been shown to accomplish long-horizon navigation in intricate, outdoor settings solely based on natural language directives. A salient feature is the lack of model fine-tuning; all three component models are trained on expansive datasets with self-supervised learning objectives and are deployed as-is. The system has demonstrated its ability to interpret and execute natural language instructions across significant distances in complex suburban terrain, while also offering disambiguation in path selection through detailed commands. Such performance metrics underscore LM-Nav's robust generalization capabilities and its proficiency in navigating complicated outdoor environments.

    b) \textbf{CLIP-NAV}~\cite{20221130CLIPNAV} introduces an innovative "zero-shot" navigation framework aimed at solving coarse-grained instruction-following tasks. The architecture is structured to dissect the guidance language into critical keyphrases, visually anchor them, and leverage the resulting grounding scores to direct the CLIP-Nav sequence-to-sequence model in predicting the agent's subsequent actions. Initially, a keyphrase extraction module isolates salient terms from the given instructions. Following this, a visual grounding module anchors these keyphrases within the environment, thereby generating a set of grounding scores. These scores serve as the basis for the sequence-to-sequence model in CLIP-Nav, which computes the next set of actions based on the agent's current state and grounding scores.

    To augment the efficacy of CLIP-Nav, the model incorporates a backtracking mechanism. This allows the agent to retrace its steps, facilitating revisions to prior decisions. Such a feature is particularly beneficial in the context of coarse-grained instruction-following tasks, wherein corrective backtracking may be essential. Additionally, the paper introduces a new performance metric, termed Relative Change in Success (RCS), to quantitatively assess generalizability in vision and language navigation tasks. 
    
  c) \textbf{SQA3D}~\cite{20230222SQA3D}, as a benchmark, aims to assess the scene comprehension ability of embodied agents.The task necessitates that the agent garner an exhaustive understanding of its orientation within a 3D environment, guided by a text-based description, and subsequently generate precise answers to questions pertaining to that understanding.

  The principal objective of SQA3D is to gauge the capacity of embodied agents to engage in logical reasoning about their immediate environment and generate answers based on such reasoning. Unlike most existing tasks, which presume that observations are made from a third-person viewpoint, SQA3D uniquely demands that agents construct and reason from an egocentric perspective of the scene.

\newsavebox{\mytikzpic}
\savebox{\mytikzpic}{%
  \begin{tikzpicture}
    \draw[->,green,line width= 2pt] (0,0) -- (0.5,0.5);
  \end{tikzpicture}%
}

    d) \textbf{ESC}~\cite{20230527ESC} (Exploration with Soft Commonsense Constrai-nts for Zero-shot Object Navigation) presents a approach to zero-shot object navigation by leveraging commonsense knowledge from pre-trained models for open-world navigation. Zhou et al. assert that effective object navigation hinges on two essential faculties: (1) Semantic scene understanding, crucial for recognizing objects and rooms, and (2) Commonsense reasoning, needed for logical inferences about probable locations of target objects based on general knowledge.

 Unlike previous methods, ESC employs pre-trained visual and language models for open-world, hint-based grounding via the GLIP model~\cite{GLIP}, enabling scene comprehension and object grounding. Large-scale image-text pre-training further facilitates generalization to new objects.
 
    Moreover, ESC deploys a pre-trained commonsense reasoning language model to deduce interrelationships between rooms and objects, applying this contextual data for spatial and object reasoning. However, translating this inferred commonsense knowledge into practical actions remains a challenge. Given that relationships between entities tend to be probabilistic rather than absolute, ESC utilizes Probabilistic Soft Logic (PSL~\cite{PSL}) to formulate "soft" commonsense constraints. These are amalgamated with conventional exploration techniques like Frontier-Based Exploration (FBE) for informed zero-shot decisions on subsequent exploration frontiers.

   e) \textbf{NavGPT}~\cite{20230529Navgpt} aims to explore the reasoning capabilities of GPT models in intricate, embodied contexts via zero-shot sequential action prediction. Zhou et al. conducted extensive experiments to show that NavGPT excels in advanced navigation planning, including the dissection of instructions into sub-goals, the incorporation of common-sense knowledge pertinent to navigation tasks, landmark identification in observed scenes, navigation progress tracking, and adjustments to plans based on unexpected developments. The study also revealed LLMs' aptitude for generating precise navigation instructions from observations and actions, as well as mapping accurate top-down metric trajectories from navigation history.

    The system accommodates multi-modal inputs, unrestricted language guidance, interactions with open-world settings, and maintains a navigation history. Utilizing GPT-4 for intricate planning, NavGPT demonstrates proficiency in sequential action prediction. It excels in subdividing instructions into sub-goals, integrating navigation-relevant common-sense knowledge, recognizing landmarks in observed settings, tracking ongoing navigation progress, and making plan adjustments based on unexpected events. Furthermore, NavGPT demonstrates spatial and historical awareness by generating trajectory instructions and plotting navigated viewpoints in an overhead view.

    Their exhaustive experimentation revealed LLMs' remarkable abilities for intricate navigation planning, including subdividing instructions into various sub-goals, incorporating relevant common-sense knowledge, recognizing landmarks in observed settings, continually monitoring navigation progress, and adjusting plans based on unforeseen occurrences. The study also confirmed that LLMs can construct navigation trajectories on metric maps and regenerate navigation instructions, thus demonstrating historical and spatial awareness in navigation assignments.

     A significant bottleneck in NavGPT is the information loss incurred when translating visual signals into natural language and condensing observations into historical records. Consequently, Zhou et al. propose future research directions that include employing LLMs with multi-modal inputs and navigation systems that leverage the advanced planning, historical, and spatial awareness capabilities of LLMs.
    \begin{figure*}[ht]
        \centering
      \includegraphics[width=\linewidth]{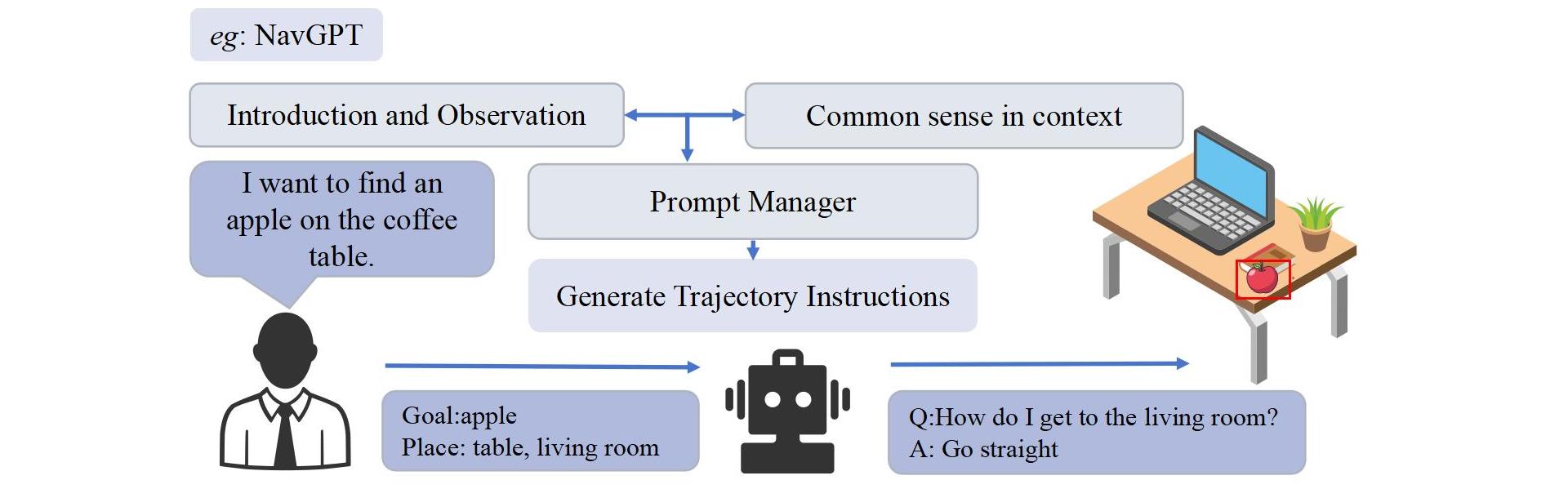}
      \caption{This figure is an example diagram for Planning.
}
      \label{fig:PlanningEGPic}
    \end{figure*}
    
    f) \textbf{VELMA}~\cite{20230712VELMA} is designed for urban VLN in Street View settings. The agent navigates based on human-generated instructions, which include landmark references and directional cues. VELMA used verbal cues, contextualized by visual observations and trajectories, to determine subsequent actions. The system identifies landmarks from human-authored navigation instructions and employs CLIP to assess their visibility in the current panoramic view, achieving a linguistic representation of visual information.

    VELMA introduces a landmark scorer to assess landmark visibility within panoramic images. This scorer calculates similarity metrics between textual descriptions of landmarks and their visual representations, using CLIP models. Each landmark receives a normalized score based on visual similarity. If this score exceeds a predefined threshold, the landmark is deemed visible.

    The landmark classification is unsupervised, given the absence of ground-truth labels for visibility. The scorer's threshold is its only tunable parameter. The agent also assesses views to its current left and right orientations. Each directional view is linked to specific textual information. Visible landmarks, along with their associated directional texts, are fed into a language expresser, which then generates verbal descriptions of the environment.These verbal observations are of two kinds: street intersections and landmarks visible within the current view. These strings encode information about the number of outgoing edges at the current node, the names of visible landmarks, and their associated directional texts.

    \begin{figure*}[ht]
        \centering
      \includegraphics[width=\linewidth]{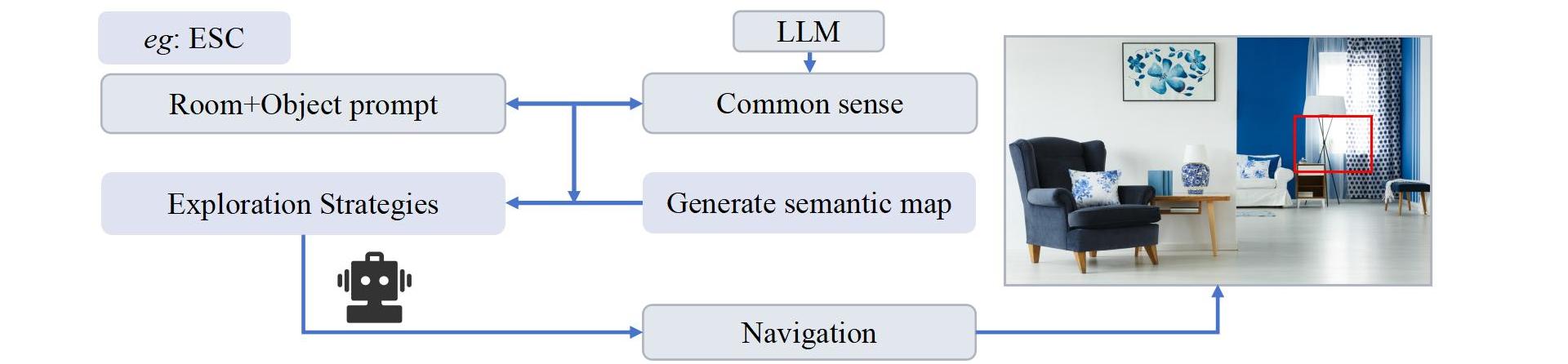}
      \caption{This figure is an example diagram for semantic understanding.
}
      \label{fig:SemanticEGPic}
    \end{figure*}

    f) \textbf{MiC}~\cite{qiao2023march} (March in Chat) is an environment-aware instruction planner that employs an LLM for dynamic dialogues, specifically designed for the REVERIE dataset. The architecture of MiC is trifurcated into three modules: Generalized Object-and-Scene-Oriented Dynamic Planning (GOSP), Step-by-step Object-and-Scene-Oriented Dynamic Planning (SODP), and Ro-om-and-Object Aware Scene Perceiver (ROASP).

    Initiating with GOSP, it queries the LLM to ascertain the target object and its probable locations, subsequently generating a rudimentary task plan. The prompt for SODP is tripartite: the first part utilizes ROASP for scene perception, acquiring room types and visible objects, and translates this information into a natural language description. The second part involves the generation of fine-grained step-by-step instructions based on the selected strategy. The final part includes previously generated instructions. These elements are concatenated and input into the LLM, which then produces detailed planning instructions for the ensuing step.

    Concerning ROASP, it not only classifies the room type but also identifies the visible objects in the agent's immediate environment. Rather than employing separate classifiers and detectors for predicting room types and visible objects, ROASP leverages the CLIP model. With CLIP's robust zero-shot image classification capabilities, ROASP efficiently performs both tasks. Specifically, it extracts room-type labels from MatterPort3D and object-type labels from REVERIE, then utilizes CLIP for feature extraction.

    In operational terms, during a task, GOSP identifies the target object by querying the LLM and utilizes the LLM’s extensive world knowledge to deduce the object's probable location. Subsequently, ROASP interprets the current scene and prompts the LLM to generate detailed step-by-step plans for the next navigation action. If ROASP discerns a room change, SODP re-queries the LLM to generate a fresh set of fine-grained instructions, aligning them with prior LLM responses. The agent then continues, following these dynamically generated instructions, executing GOSP once and iterating through ROASP and SODP until task completion.

    g) \textbf{SayNav}~\cite{20230911SayNav} is a groundbreaking framework designed to harness the common-sense knowledge encapsulated in Large Language Models (LLMs) for generalizing intricate navigation tasks in expansive and unfamiliar environments. SayNav employs a unique grounding mechanism that incrementally constructs a 3D scene graph of the explored territory, subsequently feeding this information to LLMs. This results in the generation of high-level navigation plans that are context-aware and practically implementable, executed thereafter by a pretrained low-level planner as a series of short-distance point-goal navigation sub-tasks.

    Incorporating a two-tiered planning architecture and relying on the ProcTHOR~\cite{THOR} dataset, SayNav’s high-level planner extracts subgraphs from the comprehensive 3D scene graph, concentrating on spatial relations in the immediate vicinity of the agent. These subgraphs are converted into textual prompts and presented to the LLM, which in turn produces short-term, stepwise instructions in pseudo-code format. The plans thus generated recommend efficient search strategies, ranking various locations in the room based on their likelihood of containing target objects.

    The low-level planner is tasked with generating brief motion control commands. It operates on RGBD images (320x240 resolution) and the agent's pose data, outputting basic navigational commands in alignment with standard PointNav configurations\cite{li2024curved}. To unify the operations of the high-level and low-level planners, SayNav treats each pseudo-code instruction from the LLM as a short-distance point-goal navigation sub-task. The 3D coordinates of the objects specified in each planned step determine the target points for these sub-tasks.

    SayNav distinguishes itself by its ability to dynamically generate navigation instructions and iteratively refine its future steps based on newly collected data. Performance metrics on multi-object navigation tasks demonstrate its substantial advantage over Oracle-based PointNav methods, confirming its proficiency in dynamically planning and  executing object-finding tasks in large, unfamiliar environments. Moreover, SayNav has proven its efficacy in generalizing from simulated to real-world conditions. In conclusion, SayNav emphasizes the utilization of human-like reasoning and advanced semantic understanding for achieving efficient and adaptive navigation, especially in complex or uncharted settings.
    h) \textbf{ETPNav}~\cite{an2023bevbert_cas} combines topological and metric map construction to balance long-term planning and short-term decision-making. The system processes panoramic RGB images to generate feature vectors and uses a pre-trained ViT to extract image features, while also incorporating depth information to build a metric map for real-time navigation decisions. The topological map records all observed nodes, supporting long-term planning, and represents each node's spatial position and environmental information through contextual view embeddings. The metric map is locally constructed around the current node, combining grid features from nearby visited nodes to accurately represent the spatial details of the environment for immediate navigation.
    
Additionally, the framework employs a cross-modal long-term and short-term Transformer model for both long-term planning and real-time navigation. The long-term Transformer handles long-term planning tasks by combining node embeddings and topological map encoding, while the short-term Transformer optimizes real-time decisions by integrating metric map data and current node information. This multi-modal fusion model enhances the agent's navigation ability in complex environments, enabling it to effectively perform tasks that require both visual and language understanding.

    LLMs have demonstrated significant contributions to navigation tasks, primarily through their enhanced capabilities in environmental perception and decision-making. These models integrate advanced language and image processing technologies to provide sophisticated environmental perception and decision support, crucial for tasks requiring a deep understanding of the environment and quick, accurate decision-making. Unlike traditional navigation models that are limited to predefined rules and instructions and offer limited interaction modes, LLMs excel by enabling more dynamic and flexible route planning. Researchers are continuously exploring various applications of LLMs in embodied navigation systems to further enhance their effectiveness in navigation tasks~\cite{chen2024constraint}. These applications extend beyond traditional navigation functionalities to include augmented reality perception, multimodal interaction, procedural generation, and automated decision support.

    LLMs significantly enhance the environmental perception capabilities within embodied navigation systems through their robust language and image processing abilities. This includes the recognition and interpretation of environmental features and their utilization to make optimal decisions. For example, LLMs can improve navigation performance by analyzing visual information or better understand ground information through the analysis of surface processes and interactions. In terms of decision-making, LLMs are adept at adjusting travel direction and speed based on current location and target information, enabling efficient navigation in unknown or complex environments. They adjust their route based on real-time data and target information to adapt to environmental changes and effectively complete tasks~\cite{dwivedi2022navigation}.

    For instance, the NaviLLM model~\cite{zheng2023towards} supports the execution of multiple navigation tasks such as visual, auditory, and language navigation through pattern-based instructions, enabling a unified model to perform diversely. Some studies have proposed integrating LLMs with visual encoders to enhance the model's intuitive understanding of the open world, allowing the model to process richer visual information and generate multimodal feedback. This integrated approach not only enhances the model's interactive capabilities but also improves its performance in open-world scenarios~\cite{zheng2023steve}. The powerful reasoning capabilities of LLMs significantly enhance the user experience in embodied navigation, for instance, by utilizing LLMs to interpret user commands in complex or emergency situations, ensuring that the autonomous driving system aligns with user intentions~\cite{yang2024human}. Simultaneously, by optimizing spatial perception abilities, the system provides more precise spatial location information or scene diagrams, allowing for a more accurate understanding of and response to the user's spatial position and objectives. Additionally, the DynaCon system~\cite{kim2023dynacon} utilizes LLMs for context-awareness and dynamic adaptability, enhancing the embodied navigation system's capabilities in situations requiring a deep understanding of the environment and quick, accurate decision-making. Another study~\cite{oruganti2023automating} explored the capability of LLMs to generate control scripts to support automated learning of new entries and use natural language supervision to predict the matching of labels with images, facilitating the zero-shot transfer of models to downstream tasks, thereby offering conveniences for the execution of navigation tasks.

 \subsubsection{Other Models}
 In addition to the LLM-based navigation models discussed in the previous section, this section compiles several outstanding non-LLM navigation models. These models have also achieved excellent performance through their unique approaches.
 
  a) \textbf{CoW}~\cite{20220520CoW} (CLIP on Wheels) by Gadre et al.~\cite{20220520CoW} is an innovative approach that adapts zero-shot visual models like CLIP to embodied AI tasks, particularly object navigation. The framework involves an agent identifying a target object in unfamiliar environments, defined through text. The key concept is dividing the task into zero-shot object localization and exploration.

   However, applying CLIP directly presents challenges. Firstly, CLIP struggles with precise spatial localization crucial for steering the agent. Secondly, CLIP, being static image-based, lacks mechanisms for guiding exploration. Finally, conventional fine-tuning of CLIP might reduce its robustness and generalizability.

   To address these, the CoW framework avoids fine-tuning and uses three techniques for object localization: Gradient-based (using CLIP gradients for saliency mapping), k-Patch-based (discretizing the image into sub-patches for individual CLIP model inference), and k-Language-based (matching the entire image with various captions for location information). For exploration, CoW leverages depth maps with two methods: Learning-based (incorporating a GRU, linear actor, and critic heads) and Frontier-based (a top-down map expansion approach)~\cite{yamauchi1997frontier}.

   CoW effectively repurposes zero-shot image classification models for embodied AI, achieving notable success rates by selecting appropriate methods for each task component. While user-specified target exploration is promising, real-world application remains the ultimate test.
    \begin{table*}[ht]
\caption{\textbf{Comparison of Works} This table provides a comparison of different Works, the design structure, and their respective applications. It's worth noting that each wok uses different evaluation criteria, which are not listed here. For specific details on each work, please refer to the citations at the end of the article.}
\label{tab:Benchcomp2}
\resizebox{\textwidth}{!}{
\begin{tabular}{ccccc}
\hline
\multirow{2}{*}{\textbf{Type}}                  & \multirow{2}{*}{\textbf{Works}} & \multirow{2}{*}{Multimodal}   & \multirow{2}{*}{Design Structure} & \multirow{2}{*}{Specific Applications}                           \\
                                                &                                 &                               &                                   &                                                                  \\ \hline
\multirow{10}{*}{\textbf{LLM-based Models}} & LM-Nav~\cite{20220726LMNAV}                          & \bluecheckmark & ViNG,CLIP,GPT-3                   & Translate user commands into a list of landmarks                 \\
                                                & SQA3D~\cite{20230222SQA3D}                           & \bluecheckmark & CLIP,BERT,GPT-3                   & Contextual understanding, zero-sample model exploration          \\
                                                & L3MVN~\cite{20230411L3MVN}                           & \bluecheckmark & RoBERTa                           & Provide common sense for object searching                        \\
                                                & NavGPT~\cite{20230529Navgpt}                          & \bluecheckmark & GPT-4                             & Integrate common sense knowledge and generate actions            \\
                                                & VELMA~\cite{20230712VELMA}                            & \bluecheckmark & CLIP,GPT                          & Language understanding and generate actions                      \\
                                                & CLIP-Nav~\cite{20221130CLIPNAV}                         & \bluecheckmark & CLIP,GPT-3                        & Instruction decompose                                            \\
                                                & MiC~\cite{qiao2023march}                              & \bluecheckmark & GPT                               & Planning and generate actions                                    \\
                                                & ESC~\cite{20230527ESC}                              & \bluecheckmark & Deberta                           & Soft common sense constraints and cene understanding             \\
                                                & $ A^2 $Nav~\cite{20230815A2NAV}                       & \bluecheckmark & GPT-3                             & instruction decompose                                            \\
                                                & SayNav~\cite{20230911SayNav}                           & \bluecheckmark & GPT-3.5-turbo,GPT-4               & Dynamically generate step-by-step instructions                   \\ \hline
\multirow{4}{*}{\textbf{Other Models}}           & CoW~\cite{20220520CoW}                              & \bluecheckmark & CLIP                              & Object localization,Exploring the environment                    \\
                                                & ZSON~\cite{20220624ZSON}                             & \bluecheckmark & CLIP,ResNet50                     & Generate semantic image target and                               \\
                                                & VLMAP~\cite{20230508VLMAP}                            & \bluecheckmark & LSeg,CLIP                         & Visual language pixel level embedding and 3D map location fusion \\
                                                & OVRL~\cite{20230514OVRL}                             & \bluecheckmark & ViT,LSTM                          & Process RGB observations and predict actions                     \\ \hline
\end{tabular}
}
\end{table*}

    b) \textbf{ZSON}~\cite{20220624ZSON} introduces an innovative methodology for instructing virtual robots to navigate unfamiliar terrains and identify objects without pre-existing rewards or demonstrations. Distinct from conventional ObjectNav techniques, ZSON exploits image-goal navigation (ImageNav) to transcode goal images into a multimodal, semantic embedding space. This allows for the scalable training of semantic-goal navigation (SemanticNav) agents in unannotated 3D settings. The underpinning theory of ZSON hinges on the principle of semantic similarity. By encoding goal images as semantic embeddings, the approach enables agents to navigate toward objects predicated on their semantic likeness to the goal image. This strategy assumes that objects bearing semantic resemblance to the goal image are likely located in similar spatial contexts.

    In the implementation phase, the first step entails using CLIP for pre-training to produce semantic embeddings of image targets. These embeddings encapsulate intricate semantic details about the objectives. Subsequently, a SemanticNav agent undergoes training through reinforcement learning. The agent ingests egocentric RGB observations and semantic target embeddings as input variables and utilizes a ResNet-50 encoder along with a policy network to forecast actions.In performance evaluation, the team executed extensive experiments on three ObjectNav datasets: Gibson, MP3D, and HM3D. Compared to previous zero-shot works, their method has shown a significant improvement.

    c) \textbf{VLMAP}~\cite{20230508VLMAP} presents an innovative spatial map representation that integrates pretrained visual-language features with a 3D environmental reconstruction, facilitating natural language map indexing without the need for extra labeled data. The central innovation is its capacity to intrinsically combine visual-language attributes with the 3D environmental blueprint, empowering robots to execute spatially precise navigation via natural language directives.

    The methodology for VLMAP consists of creating a spatial visual-language map capable of direct landmark or spatial reference localization via natural language. Employing readily available visual-language models and conventional 3D reconstruction libraries, VLMAP is assembled by merging visual-language characteristics with a 3D environmental model, thereby allowing natural language map indexing without supplemental labeled data.

    VLMAP advances the fields of computer vision, natural language processing, and robotics by unifying visual-language features with 3D reconstructions for precise, natural language-guided robotic navigation. The incorporation of visual-language features leverages recent progress in deep learning, while the 3D reconstruction component builds on established principles of 3D point cloud processing. The natural language capabilities are supported by an expanding corpus of research in natural language processing and human-robot interaction.Experimental evaluations of VLMAP substantiate its efficacy in enabling robots to navigate spatially precise paths through natural language commands. Tests in a simulated environment affirm the system's capability to convert natural language directives into a sequence of open-vocabulary navigation objectives that can be directly pinpointed on the map. Further, the adaptability of VLMaps for use among robots with disparate configurations to generate real-time obstacle maps is also demonstrated.

   d) \textbf{OVRL}~\cite{20230514OVRL} introduces a neural network architecture specifically aimed at visual navigation challenges. Notably, the architecture is composed of task-agnostic elements such as ViTs, convolutions, and LSTMs, obviating the need for task-specific modules. The standout feature of OVRL-V2 is a novel compression layer designed to preserve spatial details in visual navigation assignments. This layer compresses high-dimensional image attributes from the ViT model into low-dimensional features, maintaining the essential spatial context.

   The incorporation of the compression layer is grounded in the theoretical premise that retaining spatial information is vital for navigation endeavors, and this layer effectively conserves this attribute. The model further employs self-supervised learning for pretraining, which enhances its generalization performance across new environments.

   Architecturally, the paper outlines a universal agent structure consisting of a visual encoder, a goal encoder, and a recurrent policy network. The visual encoder utilizes a ViT-based module for RGB data processing. This output is then amalgamated with a goal representation and channeled through a recurrent LSTM policy network to predict actions. Training methods vary between tasks: IMAGENAV agents are trained via reinforcement learning and DD-PPO~\cite{wijmans2019dd}, while OBJECTNAV agents employ human demonstrations and behavior cloning for training. The visual encoder can either be trained de novo in an end-to-end fashion or pretrained via the MAE algorithm.

   Traditional rule-based navigation algorithm (such as Theta*~\cite{daniel2010theta}, Lifelong Planning A*~\cite{koenig2004lifelong}, Multi-Agent Path Finding (MAPF)~\cite{sharon2015conflict})s and contemporary navigation algorithms each have their strengths and weaknesses. Traditional algorithms have clear rules and stable performance, making them suitable for simple and common scenarios. They have low computational complexity, making them ideal for resource-limited devices and environments. However, the design of these algorithms' rules cannot cover all possible scenarios, leading to poor flexibility and scalability. Additionally, traditional algorithms struggle to integrate multiple data sources, limiting their ability to handle complex scenarios.

    In contrast, modern navigation algorithms, although not entirely based on large language models (LLMs), achieve multimodal data integration through the application of other visual language models. For example, by using weak rules (such as boundary exploration), these algorithms demonstrate strong adaptability, offering new ideas for non-LLM navigation models and proving effective in practical applications.

    LLM-based navigation algorithms have the capability to quickly process large amounts of real-time data and support natural language interaction, significantly improving navigation efficiency. Their adaptive and self-learning capabilities surpass those of traditional algorithms. However, LLM algorithms require substantial computational resources and rely on high-quality, extensive data. When handling user data, it is essential to ensure privacy and security to mitigate potential risks.

    In summary, traditional rule-based navigation algorithms still have advantages in specific scenarios and resource-constrained environments. However, modern navigation algorithms exhibit greater adaptability and the ability to handle complex scenarios by integrating multimodal data and applying weak rules. LLM-based navigation algorithms offer significant advantages in adaptability, natural language processing, and real-time data processing, but they require careful consideration of computational resources and data security issues.

    \subsubsection{Comparison}
    \begin{table*}[ht]
\centering
\caption{\textbf{Evaluation Metrics} 
The metrics of vision-and-language navigation~\cite{evaluation}.}
\resizebox{\linewidth}{!}{
\begin{tabular}{@{}>{\centering\arraybackslash}m{3cm} >{\centering\arraybackslash}m{7cm} >{\centering\arraybackslash}m{7cm}@{}}
\toprule[1.5pt]
\textbf{Evaluation criteria} & \textbf{Definition} & \textbf{Formula} \\
\midrule
Path length & Length of navigation trajectory from start to stop position & $\sum_{\mathbf{v}_i \in V} d(\mathbf{v}_i, \mathbf{v}_{i+1})$ \\
Navigation error & Distance between predicted path endpoint and reference path endpoint & $d(v_t, v_e)$ \\
Ideal success rate & The probability that any node in the predicted path is within a threshold distance from the endpoint of the reference path & $\mathbb{I} \left[ \left( \min_{\mathbf{v}_i \in V} d(\mathbf{v}_i, v_e) \right) \leq d_{th} \right]$ \\
Navigation success rate & The probability that the stopping position is within 3 meters of the end point of the reference path & $\mathbb{I} \left[ NE(v_t, v_e) \leq d_{th} \right]$ \\
Success rate based on path weighting & Navigation success rate based on path length weighting & $SR(v_t, v_e) \cdot \frac{1}{d_{gt}} \max \left\{ PL(V), d_{gt} \right\}$ \\
Weighted coverage score & Coverage and length scores of predicted paths relative to reference paths & $PC(P, R) \cdot LS(P, R)$ \\
Based on dynamic time warping weighted success rate & Success rate of predicted path to reference path based on dynamic time warping and spatial similarity & $SR(v_t, v_e) \cdot \exp \left( \frac{1}{|R|} \min_{w \in W} \sum_{(i_k, j_k) \in w} d(r_{i_k}, q_{j_k}) \right)$ \\
\bottomrule[1.5pt]
\end{tabular}
}
\end{table*}

    Evaluation metrics play a crucial role in assessing different models, serving as key indicators of model performance. The evaluation metrics for navigation tasks not only focus on Success Rate (SR) and Path Length (PL) but also require corresponding measures to evaluate the consistency between the navigation path and the instructions. Success-weighted Path Length (SPL) is a metric used to evaluate the quality of paths in navigation and path planning tasks. It combines the success rate of the path with the relationship between the actual path length and the ideal path length, thereby providing a comprehensive assessment of path quality. SPL reflects the efficiency and effectiveness of path planning. It considers the comparison between the actual path length and the ideal path length in navigation tasks while weighting the success rate of the path. SPL is widely used in fields such as robotic navigation, autonomous driving, and other areas involving path planning. The calculation of SPL is as follows:
    
    \[
\text{SPL} = \frac{1}{N} \sum_{i=1}^{N} S_i \frac{L_i}{\max(P_i, L_i)}
\]

    In the formula, N is the total number of navigation tasks, \(S_{i}\) is the success indicator where \(S_{i}\) = 1 if the task is successful and 
    \(S_{i}\) = 0 otherwise, \(L_{i}\) is the shortest path length from the start point to the target point for the task, and \(P_{i}\) is the actual path length for the task.
    
    We have compared the works mentioned above, evaluating aspects such as the large models employed, application domains, and additional features. These comparisons are summarized in Table \ref{tab:Benchcomp2}.At the same time, we have divided the aforementioned works into two categories: Planning, as illustrated in fig.\ref{fig:PlanningEGPic}, and Semantic Understanding, as shown in fig.\ref{fig:SemanticEGPic}. Corresponding examples are provided for each category. Furthermore, we have analyzed the performance of these works across various datasets, the details of which are provided in Tables \ref{tab:P1} and \ref{tab:P2}. It is important to clarify that comprehensive evaluation was unfeasible for some works due to the unavailability of public code. Consequently, the results are based solely on performance metrics reported in the corresponding publications. For an in-depth understanding of each work, readers are directed to the cited literature.

        \begin{table}[h]
	\centering
	\caption{Performance Comparison (Part I)}
 \label{tab:P1}
 \resizebox{0.5\textwidth}{!}{
	\begin{tabular}{ccccccc}
		\toprule
		Benchmark & \multicolumn{2}{c}{Habitat} & \multicolumn{2}{c}{RoboTHOR} & \multicolumn{2}{c}{Gibson} \\
		\cmidrule(r){2-3} \cmidrule(lr){4-5} \cmidrule(l){6-7}
		& SPL   & SR    & SPL   & SR    & SPL & SR                           \\
		\midrule
		CoW       & 6.3\% & -     & 10\%  & 16.3\% & -          & -                    \\
		ZSON      & -     & -     & -     & -      & 28\% & 36.9\%              \\	
		L3MVN     & -     & -     & -     & -      & 37.7\%    & 76.1\%         \\
		VLMAP     & 6\%   & 2.5\% & -     & -      & -           & -                   \\
		ESC       & -     & -     & 22.2\%& 38.1\% & -           & -                   \\
		\bottomrule
	\end{tabular}
 }
\end{table}
\begin{table}[ht]
	\centering
	\caption{Performance Comparison (Part II)}
  \label{tab:P2}
  \resizebox{0.5\textwidth}{!}{
	\begin{tabular}{ccccccccc}
		\toprule
		Benchmark & \multicolumn{2}{c}{HM3D}           & \multicolumn{2}{c}{MP3D}           & \multicolumn{2}{c}{REVERIE}   & \multicolumn{2}{c}{R2R}\\

  \cmidrule(r){2-3} \cmidrule(lr){4-5} \cmidrule(l){6-7} \cmidrule(l){8-9}
		& SPL   & SR                             & SPL   & SR                             & SPL  & SR       & SPL  & SR                     \\
		ZSON      & 12.6\%& 25.5\%                         & 4.8\% & 15.3\%                         & -       & -  & -    & -                   \\
		CLIP-Nav  & -     & -                              & -     & -                              & 4.56\% & 15.2\%     & - & -           \\
		L3MVN     & -     & -                              & 23.1\%& 50.4\% & -       & -  & -       & -                \\
		OVRL      & 29\%  & 64\%                           & -     & -                              & -      & -   & -    & -                   \\
		ESC       & 22\%  & 38.5\%                         & 13.7\%& 28.6\%                         & -      & -  & -       & -             \\
		NavGPT    & -     & -                              & -     & -                            & -  & -    & 29\%  & 34\%                       \\
		BEVBert  & -     & -                              & -     & -                             & -  & -   & 64\%    & 75\%                     \\
		MIC       & -     & -                              & -     & -                              & 41.97\%  & 55.74\%    & -   & -              \\
		SayNav    & 34\%& 60.32\%                           & -     & -                              & -      & -  & -  & -                        \\
		\bottomrule
	\end{tabular}
 }
\end{table}

\subsection{Datasets}
  Recently, with the accelerated advancement of AI technologies, the focus of interest has transitioned from digital interfaces to tangible environments, catalyzing the emergence of a novel research frontier—Embodied AI. As implied by its nomenclature, Embodied AI primarily investigates the dynamic interplay between an autonomous agent and its real-world environment. To execute tasks both successfully and robustly, agents require advanced proficiencies in environmental perception, encompassing, but not limited to, faculties in vision, language, reasoning, and planning. Consequently, the selection of an appropriate dataset for pre-training these agents becomes a critical determinant of their performance.

\begin{figure*}[ht]
      \centering
      \includegraphics[width=\linewidth]{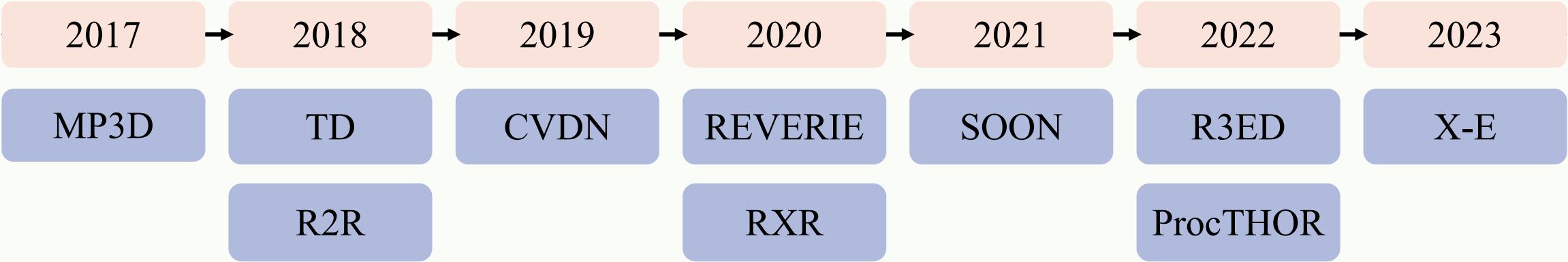}
      \caption{The timeline below provides an overview of the key, commonly referenced datasets discussed in this paper, covering a period from 2017 to 2023.
}
      \label{fig:Datasettl}
    \end{figure*}

  Current explorations within the domain of Embodied AI encompass a diverse array of applications including navigation, object manipulation, and multimodal learning. Correspondingly, the datasets requisite for these distinct research areas exhibit considerable variation. This section will present an overview of multiple datasets pertinent to embodied intelligence: MP3D~\cite{20170918MP3D}, TOUCHDOWN~\cite{2018TD}, R2R~\cite{20180405R2R}, CVDN~\cite{20191013CVDN}, REVERIE~\cite{20200106REVERIE}, RXR~\cite{20201005RXR}, SOON~\cite{20211015SOON}, ProcTHOR~\cite{THOR}, R3ED~\cite{20220703R3ED}, and X-Embodiment~\cite{openx}.  The chronological development of these datasets is depicted in Figure \ref{fig:Datasettl}.

a) \textbf{MatterPort3D (MP3D)}~\cite{20170918MP3D} is a comprehensive dataset containing 10,800 panoramic views from 194,400 RGB-D images across 90 architectural settings, focusing on 3D indoor environments. It includes RGB images, depth images, and semantic annotations for surfaces, camera poses, and segmentation, making it a key resource for self-supervised tasks and embodied AI. MP3D's detailed scenes and global perspectives make it ideal for object detection, scene segmentation, and 3D reconstruction, although its focus on indoor environments limits its broader applicability.

b) \textbf{TOUCHDOWN}~\cite{2018TD} is designed for navigation and spatial reasoning using Google Street View, containing 39,641 panoramic images from New York City. It features two tasks: Navigation and Spatial Description Resolution (SDR), emphasizing spatial reasoning and linguistic expertise. TOUCHDOWN is suitable for outdoor navigation tasks, unlike MP3D, which focuses on indoor settings, but the complexity of urban environments and inconsistent language annotations pose challenges.

c) \textbf{Room-to-Room (R2R)}~\cite{20180405R2R} is a dataset for Vision-and-Language Navigation (VLN), extending MP3D with natural language instructions. R2R focuses on navigation in indoor spaces, requiring models to understand natural language and correlate it with visual information. This dataset presents a challenging research platform for complex decision-making and long-term memory in indoor navigation.

d) \textbf{CVDN}~\cite{20191013CVDN} is a dataset of over 2,000 "human-to-human" dialogues in a photorealistic simulated domestic environment. It aims to train agents to navigate residential and commercial spaces using linguistic directives. Unlike R2R, CVDN features longer dialogues and emphasizes cooperative navigation, making it ideal for studying human-computer interaction and dynamic inquiry in navigation.

e) \textbf{REVERIE}~\cite{20200106REVERIE} includes 10,567 panoramic images and 21,702 crowd-sourced instructions for target-oriented navigation tasks. It differs from traditional VLN datasets by providing semantically rich tasks, such as "fold the towel in the bathroom with the fishing theme," which requires long-term memory and semantic understanding. REVERIE is suitable for remote entity referencing and multi-task learning.

f) \textbf{RxR}~\cite{20201005RXR} extends R2R with multilingual natural language instructions in English, German, and Spanish. It introduces spatiotemporal grounding, requiring agents to understand multilingual instructions, spatial reasoning, and long-term planning. RxR is a challenging dataset for advancing multilingual VLN.

g) \textbf{ProcTHOR}~\cite{THOR} is a dataset of 10,000 algorithmically generated 3D residential scenes designed for training embodied AI agents. It supports tasks like navigation, object manipulation, and scene understanding, and has shown robust performance across benchmarks such as Habitat 2022 and RoboTHOR challenges.

\begin{table*}[ht]
\centering
\caption{\textbf{Comparison of Datasets}. This table is a comparison among nine datasets, including seven evaluation methods.}
\label{tab:Datacomp}
\setlength{\tabcolsep}{10pt} 
\renewcommand{\arraystretch}{1.2} 

\begin{tabularx}{\textwidth}{c|ccccccccc}
\hline
\multirow{2}{*}{Dataset} &  & \multicolumn{1}{l}{}          & Size                 & \multicolumn{1}{l}{}            &  & \multicolumn{2}{c}{Content}  &  & \multirow{2}{*}{Real(Physical)} \\ \cline{3-5} \cline{7-8}
                         &  & \multicolumn{1}{l}{Panoramic} & Instruction          & \multicolumn{1}{l}{Point Cloud} &  & Human & Inclusion            &  &                                 \\ \hline
MP3D                     &  & 10,800                        & -                    & -                               &  & \bluecheckmark      & View, Annotation      &  & \bluecheckmark                            \\
TOUCHDOWN                &  & 39,641                        & -                    & -                               &  & \bluecheckmark      & Images, Edges         &  & \bluecheckmark                            \\
R2R                      &  & -                             & 21,567               & -                               &  & \bluecheckmark      & Nav-Instruction      &  & \bluecheckmark                            \\
REVERIE                  &  & 10,567                        & 21,702               & -                               &  & \bluecheckmark     & View, Instruction     &  & \bluecheckmark                            \\
RxR                      &  & -                             & 126,000              & -                               &  & \bluecheckmark     & Nav-Instruction      &  & \bluecheckmark                            \\
SOON                     &  & -                             & 3,848                & -                               &  & \bluecheckmark      & Instruction, Location &  & \bluecheckmark                            \\
ProcTHOR                 &  & 10,000                        & -                    & -                               &  & \redxmark          & View                 &  & \redxmark                                \\
R3ED                     &  & -                             & -                    & 5800                            &  & \redxmark          & PointCloud, Bbox      &  & \bluecheckmark                            \\
CVDN                     & \multicolumn{1}{c}{} & -         & -                    & -                       & \multicolumn{1}{c}{} & \multicolumn{1}{c}{\bluecheckmark} & Dialog               & \multicolumn{1}{c}{} &  \bluecheckmark                            \\ \hline
\end{tabularx}
\end{table*}

    h) \textbf{X-Embodiment}\cite{openx} is an embodied robot manipulation dataset with over one million robot trajectories from 22 instances, covering 527 skills and 160,266 tasks. It merges data from 60 pre-existing robot datasets and standardizes them into the RLDS format\cite{RLDS}, facilitating easy integration with various deep learning frameworks. This large-scale, diversified dataset is crucial for developing generalist robotic policies that can adapt to novel robots, tasks, and settings. By aggregating data across multiple platforms, X-Embodiment enables the creation of more versatile and universally applicable robotic policies, advancing the field of robot learning.

\section{Challenges and Future Directions}

LLMs face several challenges and limitations in navigation applications. Different datasets have unique strengths for various navigation and language guidance tasks. For instance, Matterport3D, R2R, RXR, and REVERIE~\cite{20170918MP3D}\cite{20180405R2R}\cite{20201005RXR}\cite{20200106REVERIE} are suitable for indoor navigation tasks, while TOUCHDOWN~\cite{2018TD} is tailored for urban environments, and ProcTHOR~\cite{THOR} caters to controlled environments. R3ED~\cite{20220703R3ED} is designed for long-term tasks with a memory module. MP3D provides high-quality RGB-D images and 3D reconstruction data but requires high hardware configurations~\cite{20170918MP3D}. TOUCHDOWN offers urban street scene data but may have inconsistent language annotations~\cite{raja2023smart}. CVDN~\cite{20191013CVDN} emphasizes cooperative navigation, with longer dialogues and path descriptions, while ProcTHOR allows for customizable virtual environments, ideal for embodied AI tasks.

LLM-based navigation models leverage powerful NLP capabilities to guide agents across environments, capable of understanding and generating natural language~\cite{qiao2023march}\cite{20230712VELMA}. However, these models typically rely on pre-trained LLMs without task-specific fine-tuning, affecting performance. Non-LLM methods, by contrast, require less data and avoid the need for extensive natural language understanding, excelling in tasks like image-text matching and visual tasks. LLMs, however, struggle with fine-grained navigation instructions and spatial reasoning, performing poorly in long-term planning and environments with traversable obstacles\cite{yan2024instrugen}\cite{mann2020language}.

Models like NaVid~\cite{zhang2024navid} integrate multimodal data (RGB video, text) to enhance navigation, with pre-trained models alleviating training costs~\cite{zhang2024mm}. Future LLM-based models may benefit from multi-sensor fusion and prediction methods to improve navigation tasks, especially in complex environments~\cite{yu2023merlin}\cite{wang2022towards}. The integration of multimodal instruction tuning can improve task performance, and privacy protection must be prioritized in the design of embodied AI systems~\cite{jordon2022synthetic}~\cite{zhang2023ethical}.

In the future, LLM-based navigation models may evolve in the following directions: 
\begin{enumerate}
\item dynamic path optimization, which is no longer bound to static space (e.g. MP3D) but moves towards dynamic space to adjust paths in real time according to dynamic targets in space. 
\item optimization of algorithms and model architectures using LLMs trained for a specific task in order to reduce resource consumption and improve efficiency.
\item integration with automated driving technologies. by fusing multimodal data, such as GPS radar, to provide high-precision navigation support while improving the user interaction experience.
\end{enumerate} 

On the whole, LLMs have great potential for development in navigation tasks, but still need to solve the challenges of data diversity, fine-grained navigation, spatial reasoning ability and interaction adaptability, and improve their application in complex environments through multimodal fusion and optimization algorithms.

\section{Conclusion}
  This paper delves into the rapidly evolving field of Embodied Intelligence, focusing on the premise that intelligence emanates from the interaction between an agent and its environment rather than being a purely internal, abstract construct. The paper scrutinizes several works in embodied navigation, delineating their variances and commonalities. Furthermore, it offers an in-depth analysis of how these works catalyze contemporary advancements in research tasks in embodied AI, shedding light on their underlying design principles and methodologies. The paper also outlines challenges and limitations inherent to LLMs in the realm of embodied intelligence. These include the absence of a direct linkage between LLMs and the physical world, the necessity for extensive training data, and the complexity of understanding and generating natural language in a diverse array of contexts. Notwithstanding these impediments, several promising research trajectories exist that could propel advancements in embodied intelligence. These encompass the formulation of more robust and adaptable language models, the investigation of innovative training methodologies and architectures, the establishment of standardized works and evaluation metrics, and the critical need for cross-disciplinary collaboration among researchers in AI, robotics, and social sciences.
\bibliographystyle{IEEEtran}
\bibliography{ref}
\nocite{*}
\end{document}